# Paralinguistic Privacy Protection at the Edge

RANYA ALOUFI, Imperial College London, UK
HAMED HADDADI, Imperial College London, UK
DAVID BOYLE, Imperial College London, UK

Voice user interfaces and digital assistants are rapidly entering our lives and becoming singular touch points spanning our devices. These *always-on* services capture and transmit our audio data to powerful cloud services for further processing and subsequent actions. Our voices and raw audio signals collected through these devices contain a host of sensitive paralinguistic information that is transmitted to service providers regardless of deliberate or false triggers. As our emotional patterns and sensitive attributes like our identity, gender, well-being, are easily inferred using deep acoustic models, we encounter a new generation of privacy risks by using these services. One approach to mitigate the risk of paralinguistic-based privacy breaches is to exploit a combination of cloud-based processing with privacy-preserving, on-device paralinguistic information learning and filtering before transmitting voice data.

In this paper we introduce *EDGY*, a configurable, lightweight, disentangled representation learning framework that transforms and filters high-dimensional voice data to identify and contain sensitive attributes at the edge prior to offloading to the cloud. We evaluate EDGY's on-device performance and explore optimization techniques, including model quantization and knowledge distillation, to enable private, accurate and efficient representation learning on resource-constrained devices. Our results show that EDGY runs in tens of milliseconds with 0.2% relative improvement in 'zero-shot' ABX score or minimal performance penalties of approximately 5.95% word error rate (WER) in learning linguistic representations from raw voice signals, using a CPU and a single-core ARM processor without specialized hardware.

## 1 INTRODUCTION

Voice user interfaces (VUIs) are commonplace for interacting with consumer IoT devices and services. VUIs use speech recognition technology to enable seamless interaction between users and their devices. For example, smart assistants (e.g., Google Assistant, Amazon Echo, and Apple Siri) and voice services (e.g., Google Search) use VUIs to activate a voice assistant to trigger actions on IoT devices, or perform tasks such as browsing the Internet and/or reading news and playing music. The majority of these voice-controlled devices are triggered with some *wake word* or activation phrase like 'Okay, Google', 'Alexa', or 'Hey, Siri', to inform the system that speech-based data will be received. We also know that they all suffer from frequent false activations [22]. Once a voice stream is captured by a device, analysis is outsourced to the provider's cloud services that perform automatic speech recognition (ASR), speaker verification (SV), and natural language processing (NLP). This frequently involves communicating instructions to other connected devices, appliances, and third-party systems. Finally, text-to-speech services are often employed to speak back to the user. This is shown in Figure 1 (A). While VUIs are offering new levels of convenience and changing the user experience, these technologies raise new and important security and privacy concerns. The voice signal contains linguistic and paralinguistic information, where the latter is rich with interpretable information such as age, gender, and health status [70]. Paralinguistic information can therefore be considered as a rich source of personal and sensitive data. Our voice also contains indicators of our mood, emotions, physical and mental well-being, and thus raises unprecedented security and privacy concerns where raw or inferred data may be used to manipulate us and/or shared with third parties. Various neural network architectures, such as autoencoder networks (AE) and convolutional neural networks (CNN), have been proposed to tackle a diverse set of linguistics (e.g., speech recognition [83]) and computational paralinguistics applications (e.g., speaker recognition [85], emotion recognition [46], and detecting COVID-19 symptoms [21]). For example, recent end-to-end (E2E) automatic speech recognition systems rely on an autoencoder architecture so as to simplify the traditional ASR system into a single neural network [15, 18, 83]. These ASR models use an encoder to encode the input acoustic feature sequence into a vector, which encapsulates the input speech information to help the decoder in predicting the sequence of symbols, as shown in Figure 1(B). Cummins *et al.*, in [20], perform speech-based health analysis using Deep Learning (DL) approaches for early diagnosis of conditions including physical and cognitive load and Parkinson's disease. Although these deep models have comparable performance with more conventional approaches like Hidden Markov model (HMM) based ASR, they have been designed without considering potential privacy vulnerabilities given the need to train on real voice data.

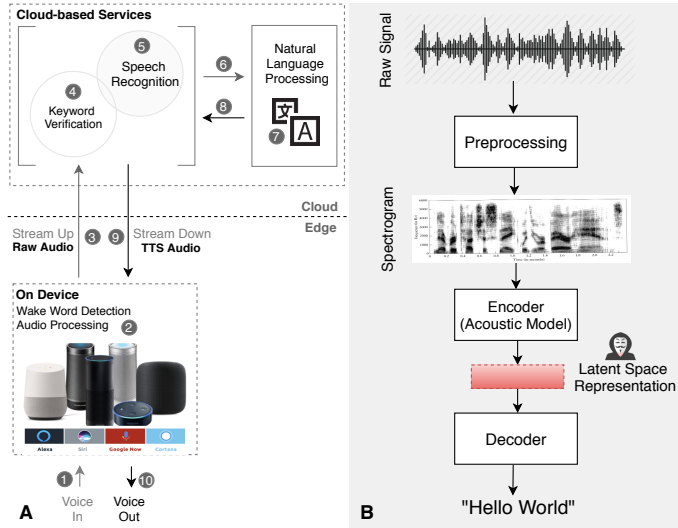

Fig. 1. (A) Voice-controlled Systems, and (B) End-to-End Automatic Speech Recognition Systems (E2E ASR).

In this paper, we present *EDGY*, a hybrid privacy-preservation approach incorporating on-device paralinguistic information filtering with cloud-based processing. EDGY pursues two design principles: it enables *primary tasks* such as speech recognition, while removing sensitive attributes in the raw voice data before sharing it with the service provider. The first is a collaborative *edge-cloud* architecture [61, 82], which adaptively partitions DL computation between edge devices (for privacy preservation) and the cloud server (for generous processing capabilities and storage). Computational partitioning alone, however, is not enough to satisfy privacy-preservation requirements [72], and we therefore leverage *disentangled representation learning* to explicitly learn independent factors in the raw data [40]. EDGY further combines DL partitioning with optimization techniques to accelerate inference at the edge. Our prototype implementation and extensive evaluations are performed using a Raspberry Pi 4 and MacBook Pro i7 as example edge and server devices to demonstrate EDGY's effectiveness in running in tens of milliseconds at the edge and an upper bound of a few seconds for the overall computation, with minimal performance penalties or accuracy losses in the tasks of interest.

In summary, our contributions are:
- We propose EDGY, a hybrid privacy-preserving approach to delivering voice-based services that incorporate on-device paralinguistic information filtering with cloud-based processing. Filtration of the voice data is based on disentangled representation learning, building on our prior work in [3, 4]. We show that the disentanglement can strengthen the edge deployment

- and can be leveraged as a critical step in developing future applications for privacy-preserving voice analytics.
- We build EDGY as a composable system to enable *configurable privacy* as well as facilitate its deployment on embedded/mobile devices. We demonstrate that a collaborative 'edge-cloud' architecture with DNN optimization techniques can effectively accelerate inference at the edge in tens of milliseconds with insignificant accuracy losses in the tasks of interest.
- We adopt 'zero-shot' linguistic metrics [57] and word error rate (WER) to evaluate the linguistic representations at two levels (acoustic and language), and through two modes of operation, (encoding and generation tasks), while using the accuracy metric to estimate the paralinguistic information. The generation task is an additional step to compare EDGY performance with current state-of-the-art ASR services and to allow practical integration with current cloud-based models.
- We experimentally evaluate the proposed framework over various datasets and run a systematic analysis of its performance at the edge under different privacy configurations, and the results show its effectiveness in learning linguistic representations (encoding task) with 0.2% relative improvement in 'zero-shot' ABX score or minimal performance penalties of 5.95% WER compared with Amazon Transcript cloud-based service and a local transcription model (generation task). Hence, such representations are confronting privacy leakage by filtering sensitive attributes with a classification accuracy drop to 34%-58% (*i.e.,* over multi/binary attributes). Our code is openly available online[1].

The paper is organized into seven sections. Following the introduction, we provide a general background about model optimization techniques for running DL at the edge and existing edge-based speech processing works in Section 2. We formulate the threat model and a propose the EDGY defense framework in Section 3. Section 4 presents the experimental settings and the implemented model optimization techniques. We evaluate our experimental results in Section 5 before providing discussion and highlighting directions for future work in Section 6. We conclude the paper in Section 7.

## 2 BACKGROUND AND RELATED WORK

### 2.1 Deep Learning at the Edge

**Edge Computing for Privacy-Preserving DL.** Edge computing is increasingly adopted in IoT systems to improve user experience and improve individuals' privacy [61, 82]. Running deep models on edge devices in practice presents several challenges. These challenges include: (i) the need to maintain high prediction accuracy with a low latency [29], and (ii) executing within the limitations of the resources available, such as memory and processing capacity of embedded devices considering the conventionally high computational requirements of these models.

DL models are generally deployed in the cloud while edge devices merely collect and send raw data to cloud-based services and receive the DL inference results. Cloud-only inference, however, risks privacy violations (i.e., inference of sensitive information). To address this issue, researchers have proposed edge computing techniques. Often, DL models must be further optimized to fit and run efficiently on resource-constrained edge devices, while carefully managing the trade-off between inference accuracy and execution time. Partitioning large DL models across mobile devices and cloud servers is an appealing solution that filters data locally before sending it to the cloud. This approach may be used to protect users' privacy by ensuring that sensitive data is not unnecessarily transmitted to service providers. In [61], a hybrid framework for privacy-preserving analytics is presented by splitting a deep neural network into a feature extractor module on the user side, and a classifier module on the cloud side.

---

[1] https://github.com/RanyaJumah/EDGY

Although most existing work in the area has looked at signals from inertial measurement unit (IMU) sensors, typically recorded while the user is performing different activities [12, 49–51, 68], we demonstrate that using the encoder part of an autoencoder to sanitize data can also be used for privacy protection in the context of speech. Since breaking the computation between the edge and cloud is not sufficient for privacy preservation purposes, we strengthen this method by learning disentangled representations in the raw data at the edge, and then filtering sensitive information before sharing it with cloud-based services.

**Optimizing DL Models for the Edge.** DNN models are usually computationally intensive and have high memory requirements, making them difficult to deploy on many IoT devices. For example, recurrent neural networks (RNNs) which are commonly used in applications such as speech processing, time-series analysis, and natural language processing, can be large and compute-intensive due to a large number of model parameters (e.g., 67 million for bidirectional RNNs) [55]. This makes it difficult to deploy these models on resource-constrained devices. Optimizing DL models by quantizing their weights can, however, reduce these resource requirements. Narang *et al.* [55] propose a method to reduce the model weights in RNNs to deploy these models efficiently on embedded or mobile devices. Similarly, Thakker *et al.* [76] significantly compress RNNs without negatively impacting task accuracy using Kronecker products (KP) to quantize the resulting models to 8-bits.

Optimizing the neural network architecture using quantization and pruning can lead to significant efficiency improvement in many speech processing applications. For example, He *et al.* [29] propose an end-to-end speech recognizer for on-device applications such as voice commands and voice search which runs twice as fast as real-time on a Google Pixel phone. They do this by quantizing model parameters from 32-bit floating-point to 8-bit fixed-point precision to reduce memory footprint and speed up computation. Zhai *et al.* [87] proposed SqueezeWave, a lightweight flow-based vocoder. SqueezeWave aims to address the expensive computational cost required by the real-time speech synthesis task. The proposed vocoder translates intermediate acoustic features into an audio waveform on a MacBook Pro and Raspberry Pi 4B to generate a high-quality speech. The most important challenge, however, is to ensure that there is no significant loss in terms of model accuracy after being optimized. Specifically, we analyze different optimization techniques (e.g., filter-pruning, weight-pruning, and quantization) to fulfill our goal in learning privacy-preserving representation from the raw data in near real-time with as little cost as possible to model performance.

## 2.2 Deep Representation Learning

**Disentangled Representation.** Learning speech representations that are invariant to differences in speakers, language, environments, microphones, etc., is incredibly challenging [47]. To address this challenge, numerous variants of Variational Autoencoders (VAEs) have been proposed to learn robust disentangled representations due to their generative nature and distribution learning abilities. Hsu *et al.* in [32] propose the Factorized Hierarchical VAE (FHVAE) model to learn hierarchical representation in sequential data such as speech at different time scales. Their model aims to separate between sequence-level and segment-level attributes to capture multi-scale factors in an unsupervised manner. There is an extended trend towards learning disentangled representations in the speech domain as they promise to enhance robustness, interpretability, and generalization to unseen examples on downstream tasks. The overall goal of disentangling is to improve the quality of the latent representations by explicitly separating the underlying factors of the observed data [40]. Speech signals simultaneously encode linguistically relevant information, e.g. phoneme, and linguistically irrelevant information, i.e., paralinguistic information. In the case of speech processing, an ideal disentangled representation would be able to separate fine-grained factors such as speaker identity, noise, recording channels, and prosody, as well as the linguistic content [24]. Thus, disentanglement will allow learning of salient and robust representations from the speech that are essential for applications including speech recognition [63], prosody transfer [75, 89], speaker verification [65], speech synthesis [33, 75], and voice conversion [34], among other applications. Although the focus of these works is to raise the efficiency and effectiveness of speech processing applications (e.g. speech recognition, speaker verification, and language translation), in this paper we highlight the benefit of *learning disentangled representation* to learn privacy-preserving speech representations, as well as showing how disentanglement can be useful in transparently protecting user privacy.

**Privacy-preserving Speech Representation.** Learning privacy-preserving representations in speech data is relatively unexplored [47]. Aloufi *et al.* [4] investigate the scenario whereby attackers can infer a significant amount of private information by observing the output of state-of-art underlying deep acoustic models for speech processing tasks. In [56], Nautsch *et al.* demonstrate the importance of the development of privacy-preserving technologies to protect speech signals and highlight the importance of applying these technologies to protect speakers and speech characterization in recordings.

The recent VoicePrivacy initiative [77] promotes the development of anonymization methods that aim to suppress personally identifiable information in speech (i.e., speaker identity) while leaving other attributes such as linguistic content intact. Most of the proposed works focus on protecting/anonymizing the speaker identity using voice conversion (VC) mechanisms [2, 66, 73, 74]. VoiceMask, for example, was proposed to mitigate the security and privacy risks of voice input on mobile devices by concealing voiceprints [66]. It aims to strengthen users' identity privacy by sanitizing the voice signal received from the microphone and then sending the perturbed speech to the voice input apps or the cloud. However, these VC methods aim to protect speaker identity against different leakage attacks depending on the attacker's knowledge of the anonymization method (i.e., ignorant, informed, and semi-informed) [44]. They found that when the attacker has complete knowledge of the VC scheme and target speaker mapping, none of the existing VC methods will be able to protect the speaker identity. Thus, disentangled-based VC might strengthen speaker identity protection by avoiding the leakage of private speaker attributes into the content embeddings. Similar to our work, Srivastava *et al.* in [73] proposed an on-device encoder to protect the speaker identity using adversarial training to learn representations that perform well in ASR while hiding speaker identity. They conclude that the adversarial training does not immediately generalize to produce anonymous representations in speech (i.e., that could be limited by the size of the training set). In [26], the authors combine different federated learning and differential privacy mechanisms to improve on-device speaker verification while protecting user privacy. Beside speaker identity, various works have been proposed to protect speaker gender [36] and emotion [3]. In [3], an edge-based system is proposed to filter affect patterns from a user's voice before sharing it with cloud services for further analysis.

Considering the 'configurable privacy' principle, we assume that privacy is subjective, with varying sensitivity between users which may even depend on the services with which these systems communicate. For example, in [88], PDVocal is proposed as a privacy-preserving and passive-sensing system to enable monitoring and estimating the risk of Parkinson's disease in daily life. Unlike other approaches, however, we seek to protect the privacy of multiple user attributes for IoT scenarios that depend on voice input or speech analysis, i.e., sanitizing the speech signal of attributes a user may not wish to share, but without adversely affecting the functionality or experience. We also emphasize the importance of learning disentangled speech representation for optimizing the privacy-utility trade-off and transparently promoting privacy.

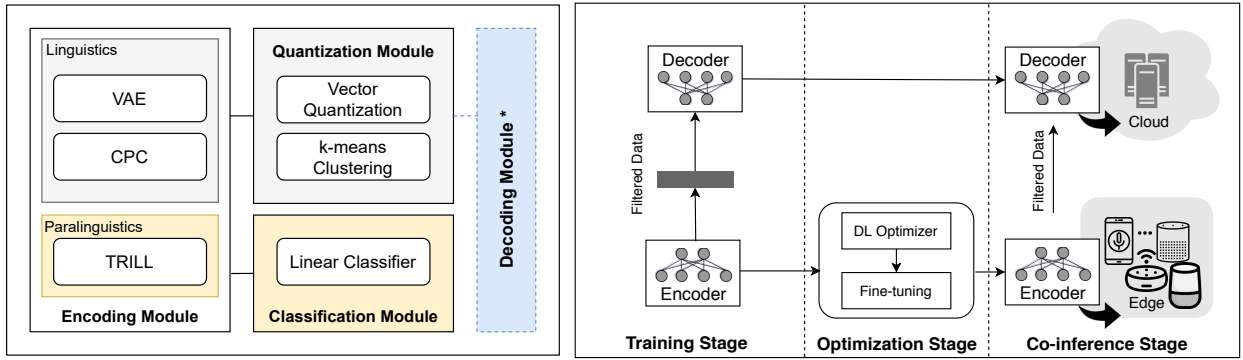

Fig. 2. The basic modules of EDGY: encoding, quantization, and classification. Each module learns a distinct type of information.

Fig. 3. A hybrid 'edge-cloud' architecture for privacy protection at the edge; starting by training the model, then partitioning and optimizing it for further edge deployment.

## 3 EDGY
### 3.1 Threat Model

We consider an adversary with full access to user data with the aim to correctly infer sensitive attributes (e.g., gender, emotion, and health status) about users by exploiting a secondary use of the same data collected for the main task. Specifically, the attacker could be any party (e.g., a service provider, advertiser, data broker, or a surveillance agency) with interest in users' sensitive attributes. The service providers could use these attributes for targeting content, or data brokers might profit from selling these data to other parties like advertisers or insurance companies, while surveillance agencies may use these attributes to recognize and track activities and behaviors. In the settings of the current system, all VUI providers have access to raw data that contains all the paralinguistic information needed to infer myriad sensitive attributes, including emotions [54], age, gender, personality, friendliness, mood, and mental health. For instance, Amazon has patented technology that can analyze users' voices to determine emotions and/or mental health conditions. This allows understanding speaker commands and responding according to their feeling to provide highly personalized content [37]. We consider *always-on* IoT services (*e.g.,* voice assistants (VA) like Amazon's Alexa and Google Home), thus, after the activation step, these services capture and transmit the speech data to powerful cloud services for further processing and subsequent actions. Our work is designed to protect the sensitive attributes contained in shared speech data from potential inference attacks (*i.e.,* honest-but-curious attacks), while adversarial examples/spoofing attacks are beyond the scope of this work. We strive to protect users from the multi-purpose inferences of authorities who, besides providing legitimate services, may attempt to infer additional sensitive information, *e.g.,* user emotion, tone, gender, age, ethnicity, etc.

**An Open Inference Attack Vector.** It is possible to accurately infer a user's sensitive and private attributes (*e.g.,* their gender, emotion, or health status) from deep acoustic models (*e.g.,* DeepSpeech2 [6]). An attacker (*e.g.,* a 'curious' service provider) may use an acoustic model trained for speech recognition or speaker verification to learn further sensitive attributes from user input even if not present in its training data. To investigate the effectiveness of such an attack, we used the output of the DeepSpeech2 model and attach different classifiers (*i.e.,* emotion and gender recognition) to demonstrate the potential privacy leakage caused by these deep acoustic models based on each of the datasets described in Section 4. We measured an attack's success as the increase in inference accuracy over random guessing [86]. We found that a relatively weak attacker (*i.e.,* using methods that include logistic regression, random forest, multi-layer perceptron, and support vector machine classifiers) can achieve high accuracy in inferring sensitive attributes, ranging from 40% to 99.4%, *i.e.,* significantly better than guessing at random.

## 3.2 Design Overview

Speech communication in human interaction can broadly convey information on two layers: (1) a linguistic layer that refers to the meaningful units of information structure in the speech signal, including phonemes (i.e., the smallest speech unit that may cause a change of meaning within a language [31]), words, phrases, and sentences, and (2) a paralinguistic/extralinguistic layer that refers to non-verbal phenomena, including speaker traits and states [70]. The phonetic content affects the segment level, while the speaker characteristic affects the sequence level [32]. Thus, the speech signal could be disentangled into several independent factors, each of which carries a different type of information.

**Core Design**: In our context, the idea is to *disentangle* the factors related to the task we want to compute per layer. We aim to demonstrate the effectiveness of *learning disentangled representations* in preserving the sensitive attributes in the user data. Such disentanglement can be beneficial to enable decentralized privacy-aware analytics and promote transparency in protecting users' privacy.

*3.2.1 Linguistics Layer.* Discrete units (e.g., phonemes) highlight linguistically relevant representations of the speech signal in a highly compact format [9, 57, 69, 78], while being invariant to speaker-specific and background noise details. These representations can be used to bootstrap training in speech systems and reduce requirements on labeled data for zero-resource languages [80]. It can also enable privacy-preserving paralinguistics.

Disentangle-based learning techniques prevent the speaker information from leaking into the content embeddings either by reducing the dimension or quantizing the content embedding as a powerful information bottleneck [19]. These techniques include: propagating reversed gradient from the speaker classifier [17], applying instance normalization [16], and quantizing the representation [78]. Thus, to achieve our goal in *learning disentangled representations* for preserving privacy, we investigate clustering approaches (e.g., $k$-means and vector quantization (VQ)) as an information bottleneck, and propose three models: $k$-means and Vector-quantized Contrastive Predictive Coding ($k$-means/VQ- CPC) and Vector-quantized Variational Autoencoder (VQ-VAE) to extract the phonetic content, while being invariant to low-level information.

One motivation to apply clustering approaches is that implementing clustering/quantization can capture high-level semantic content from the speech signal, e.g., phoneme due to the discrete nature of phonetic units [48]. The input speech sequence $x$ is first encoded into the frame-level continuous vector with length $t$. Then, the quantize layer projects the latent representation from the encoder module into the closet point in the codebook by selecting one entry from a fixed-size codebook $c = [c_1, c_2, ..., c_k]$, where $k$ is the size of the codebook.

**$K$-means/Vector-quantized- CPC.** CPC [60] is a self-supervised learning method that learns representations from a sequence by trying to predict future observations with a contrastive loss. Given an input signal $x$, the CPC model embeds $x$ to a sequence of embeddings $z = (z_1, \ldots, z_t)$ at a given rate using a non-linear encoder $E$. At each time step $t$, the autoregressive model $G$ takes as input the available embeddings $z_1, \ldots, z_t$ and produces a context latent representation $c_t = G(z_1, ..., z_t)$. Given the context $c_t$, the CPC model tries to predict the $K$ next future embeddings $\{z_{t+k}\}\ 1 \leq k \leq K$ by minimizing the following constrastive loss:

$$\mathcal{L}_t = -\frac{1}{K} \sum_{k=1}^{k} \log \left[ \frac{\exp\left(z_t^\top +_k W_k c_t\right)}{\sum_{\bar{z} \in \mathcal{N}_t} \exp\left(\bar{z}^\top W_k c_t\right)} \right] \quad (1)$$

where $\mathcal{N}_t$ is a random subset of negative embedding samples, and $W_k$ is a linear classifier used to predict the future $k$-step observation.

Rozé et al. [57] train a $k$-means clustering module on the outputs of the CPC model. After training the $k$-means clustering, the continuous features are assigned to a cluster, and the input speech sequence $x$ can then be discretized to a sequence of discrete units corresponding to the assigned clusters. Similarly, VQ-CPC [79] incorporates vector quantization with the CPC model to discretize the continuous features and capture phonetic contrasts.

**Vector-quantized- VAE.** VQ-VAE [78] model uses a Vector Quantization (VQ) technique to produce the discrete latent space. During the forward pass, the output of the encoder $z_e(x)$ is mapped to the closest entry $c_i$ in a discrete codebook of $c = [c_1, c_2, ..., c_k]$. Precisely, VQ-VAE finds the nearest codebook using Eq.1 and uses it as the quantized representation $z_q(x) = c_q(x)$ which is passed to the decoder as content information.

$$q(x) = \text{argmin}_i \|z_e(x) - c_i\|_2^2 \quad (2)$$

The transition from $z_e(x)$ to $z_q(x)$ does not allow gradient back-propagation due to the argmin function, but uses a straight-through estimator [10]. VQ-VAE is trained using a sum of three loss terms (in Eq. 2): the negative log-likelihood of the reconstruction, which uses the straight-through estimator to bring the gradient from the decoder to the encoder, and two VQ-related terms: the distance from each prototype to its assigned vectors and the commitment cost [78].

$$L = \log p(x|z_q(x)) + \|sg[z_e(x)] - c_q(x)\|_2^2 + \beta \|z_e(x) - sg[c_q(x)]\|_2^2 \quad (3)$$

Note sg($\cdot$) denotes the stop-gradient operation that zeros the gradient with respect to its argument during the backward pass. For a more details refer to [78].

*3.2.2 Paralinguistic Layer.* Information on speaker characteristics can be useful for various paralinguistics tasks such as speaker authentication [26] or Parkinson's disease detection [88], however, such information is often private. Thus, personalization and on-device training are increasingly important for these tasks, since performing computations locally can improve both privacy and latency [26].

Non-semantic aspects of the speech signal (e.g., speaker identity, language, and emotional state) generally change more slowly than the phonetic and lexical aspects used to explicitly convey meaning (i.e., ASR) [71]. Following [64, 71], we use TRIpLet Loss network (TRILL) to learn speaker-specific embedding that can be adapted to a variety of downstream paralinguistic tasks such as speech emotion recognition, speaker identification, language identification, and medical diagnosis. TRILL representation is trained using a self-supervised approach and uses triplet loss-based metric learning, assuming that segments closer in time are also closer in the embedding space. Formally, a large collection of example triplets of the form $z = (x_i, x_j, x_k)$ (namely anchor, positive, and negative examples) is sampled from unlabeled speech collection represented as a sequence $X = x_1 x_2 ... x_N$ [71]. The distance from the baseline (anchor) example to the positive (truth) example is minimized, and the distance from the baseline (anchor) example to the negative (false) example is maximized. The loss incurred by each triplet is then given by:

$$\mathcal{L}(z) = \sum_{i=1}^{N} \left[ \|g(x_i) - g(x_j)\|_2^2 - \|g(x_i) - g(x_k)\|_2^2 + \delta \right]_+ \quad (4)$$

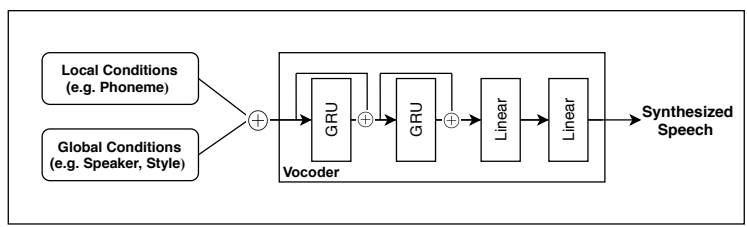

Fig. 4. An overview of the Vocoder's workflow: it concatenates a global (sequence) and local (segment) to reconstruct the output (WaveRNN [38])

where $\|.\|_2$ is the $L_2$ norm, $[.]_+$ is standard hinge loss, and $\delta$ is a nonnegative margin hyperparameter.

To support our goal of achieving configurable privacy in processing and sharing speech data, learning these generally-useful paralinguistics representations will help to enable personalization as well as serve as a first step toward achieving paralinguistics privacy in a decentralized manner, as proposed in [26].

*3.2.3 Reconstruction.* Speech reconstruction/generation is often implemented in the form of autoencoding, where speech is first encoded into a low-dimensional space, and then decoded back to speech. In the speech domain, a *vocoder* acts as a decoder and learns to reconstruct audio waveforms from acoustic features [59]. For example, Wave Recurrent Neural Networks (WaveRNN) [38] uses linear prediction with recurrent neural networks to synthesize neural audio. Specifically, it will combine the speaker identity (i.e., global condition) and linguistic context (i.e., local condition) to generate the synthesized speech, as shown in Figure 4. Thus, discovering informative discrete speech units from raw audio (i.e., zero-shot fashion) opens up the possibility of addressing the effect of non-linguistic variability (e.g., channel and speaker identity) on the linguistic quality of the reconstructed speech. It is also allow us to evaluate speech generation (i.e., speech content) at many linguistic levels.

## 3.3 Optimizing for Edge-friendly Inference

Inference efficiency is a significant challenge when deploying DL models at the edge given restrictions on processing, memory, and in some cases, power consumption. To address this challenge, we explore a variety of techniques that involve reducing model parameters with pruning and/or reducing representational precision with quantization to support efficient inference at the edge. We elaborate in the following:

*3.3.1 Quantization.* Quantized models are those where we represent the models with lower precision [35]. The main feature of existing quantization frameworks is usually the ability to quantize the weights and/or activations of the model from 32-bit floating-point into lower bit-width representations without sacrificing much of the model accuracy.

Quantization is motivated by the deployment of ML models in resource-constrained environments like mobile phones or embedded devices. For example, fixed-point quantization approaches can be implemented on the weight and activation to reduce resource consumption on devices. We thus experiment with the quantization technique on the model parameters to discover its effect on the compression of the model and increased speed of inference time at the edge with minimal detriment to its prediction accuracy.

*3.3.2 Knowledge Distillation.* Knowledge distillation refers to the idea of model compression where a complex model (i.e., teacher model) will be used to distill its knowledge to the small model (i.e., student model) without a significant drop in the prediction accuracy [13, 30].

In this process, the teacher network or an ensemble model can extract important features from the given data and produce better predictions. Then the student network with the supervision of the teacher model will be able to produce comparable results. Thus, this distillation compresses the knowledge in an ensemble model into a single model which is much easier to deploy on embedded devices. Following [64], we turn to knowledge distillation where a generally-useful speech representation model will be used to distill its knowledge to the small model, which is fast enough for on-device applications.

## 3.4 Hybrid Deployment

EDGY is designed as a composable system to enable 'configurable privacy' in various deployment environments with resource constrained devices. It currently consists of three basic modules: encoding, quantization, and classification. It is possible to add fourth unit, decoding (if needed according to the application context), as shown in Figure 2.

Assume that the processing pipeline starts with the encoding module. The encoding module consists of various models each trained on public data using an unsupervised learning approach. The module learns information over two different layers (i.e., linguistics and paralinguistics), and after verifying the ability of the model to extract useful acoustics representations, the following are the possible application scenarios under the collaborative "edge-cloud" paradigm: we suggest partitioning the encoding with the quantization module and/or the encoding with the classification module and deploying it on the resource-constrained edge devices, while the decoding module, if implemented, can be performed in the cloud; see Figure 3. For linguistics embeddings, there have been recent attempts to recognize speech through the direct use of the quantized representations (i.e., speech-related embedding) in NLP algorithms without the need for decoding these representations. For example, after using vector-quantized/clustering modules to quantize the dense representations from the speech segments, well-performing NLP algorithms (e.g. BERT) were then applied to these quantized representations, which achieve promising state-of-the-art results in phoneme classification and speech recognition [8]. In addition, paralinguistics embeddings can be used separately for further local authentication or personalization purposes.

The decoding part may be implemented for the purpose of generating voice. It is reasonable to assume that service providers may want to keep recordings in user records for the sake of transparency (*e.g.,* to facilitate GDPR). We assume that service providers may want to offer personalized services to their users. Using the proposed framework, therefore, and by decomposing the processing between the edge and the cloud, can help to achieve this objective in a privacy-preserving manner. More precisely, learning disentangled representations by the proposed framework at the edge will allow more control over the sharing of these representations. For example, service providers may train a decoder using the speaker embedding when using the service for the first time, and then the encoder at the edge sends only the speech-related embedding to regenerate the user recordings.

## 4 EXPERIMENTAL FRAMEWORK

In this section, we briefly describe the datasets used (LibriSpeech, VoxCeleb, CREMA-D, SAVEE, and Common Voice) and our experimental setup, highlighting baseline settings as well as optimization techniques used to improve EDGY's performance. To support reproducibility, our code and research artefacts are publicly available in addition to instructions about how to reproduce the implementation/results[1].

### 4.1 Datasets

We use a number of real-world datasets that were recorded for various purposes including speech recognition, speaker recognition, accents, and emotion recognition, to train EDGY and examine its effectiveness in protecting paralinguistic information. The details of each dataset are as follows:
**LibriSpeech.** LibriSpeech [62] is a large dataset of approximately 1,000 hours of reading of English.

Table 1. Popular smart speakers specifications

| Category | Device/Service | Processor Type | RAM | Storage Memory |
|---|---|---|---|---|
| Smart Speaker | Amazon Echo | ARM Cortex-A8 | 256 MB | 4 GB |
| | Google Home | ARM Cortex-A9 | 512 MB | 2 GB |
| | Apple HomePod | Apple A8 | 1 GB | N/A |

Table 2. Datasets for paralinguistics downstream tasks

| Dataset | Target | No. Classes | No. Samples | No. Speakers |
|---|---|---|---|---|
| VoxCeleb | Speaker identification | 1,251 | 153,514 | 1,251 |
| CREMA-D | Emotion | 6 | 7,442 | 91 |
| SAVEE | Emotion | 7 | 480 | 4 |
| Common Voice | Accents | 17 | 66,173 | 67,608 |

It was derived from reading audiobooks from the LibriVox project, and was recorded to facilitate the development of automatic speech recognition systems. We use the train-clean100 set and test set.

**VoxCeleb.** The VoxCeleb dataset [53] contains over 100,000 utterances from 1,251 celebrities, extracted from videos uploaded to YouTube. It was curated to facilitate the development of automatic speaker recognition systems. We use the VoxCeleb1 subset of about 1,200 recordings.

**CREMA-D.** Crowd Sourced Emotional Multimodal Actors Dataset [14] is an audio-visual data set for emotion recognition containing 7,442 recordings for 91 actors (i.e., 48 male and 43 female) with diverse ethnic backgrounds (i.e., African America, Asian, Caucasian, Hispanic, and Unspecified). It was recorded to facilitate the development of multimodal emotion recognition systems. It includes seven emotions: calm, happy, sad, angry, fear, surprise, and disgust, as well as neutral expression. We use the entire dataset.

**SAVEE.** Surrey Audio-Visual Expressed Emotion database [28]. It consists of phonetically-balanced sentences from standard TIMIT (acoustic-phonetic continuous speech dataset) uttered by four English actors with a total size of 480 utterances. It was primarily recorded to facilitate the development of multimodal emotion recognition systems. It contains expressions of seven emotions: calm, happy, sad, angry, fearful, surprise, and disgust, as well as neutral. We use the entire dataset.

**Common Voice.** Common Voice Dataset [7] is a massively multilingual collection of transcribed speech collected via Mozilla's Common Voice initiative for speech technology research and development. The dataset currently consists of 7,335 validated hours in 60 languages including demographic metadata like age, gender, and accent per recording. We use the entire English set.

**Preprocessing.** In our experiments, we use a sampling rate of 16 Hz for the recordings. We extract 80-bins Mel-spectrograms using libros v0.7.2, with an FFT size of 2048, hop size 256, and window size 600 as Mel-spectrogram parameters.

## 4.2 Experimental Setting

We conduct our experiments on a Z8 G4 workstation with Intel (R) Xeon (R) Gold 6148 (2.8 GHz) CPU and 256 GB RAM. The operating system is Ubuntu 18.04. We train and fine-tune all models on an NVIDIA Quadro RTX 5000 GPU. We run a total of 50 experiments divided into 18 experiments with various compression techniques, followed by 32 experiments for privacy estimation. Then, we deploy the EDGY on a MacBook Pro with an Quad-Core Intel i7 CPU and a Raspberry Pi 4B with a Broadcom BCM2711 CPU, quad core Cortex-A72 (ARM v8) 64-bit, which is similar to the specifications of current voice-controlled devices, some examples of which are shown in Table 1.

*4.2.1 Model Architecture.* For the linguistics embedding, the target is to learn discrete units (i.e., speaker-invariant) useful for speech recognition and phoneme classification. We apply three different vector quantized-based models which are: CPC-kmean clustering [57], CPC-VQ [79], and VAE-VQ [78] to extract the phonetic content. All of these models start by encoding the audio signal at the encoder, then the encoder output (latent vectors) passes through a vector quantization layer to become a sequence of quantized representation that serves as the speech embedding.

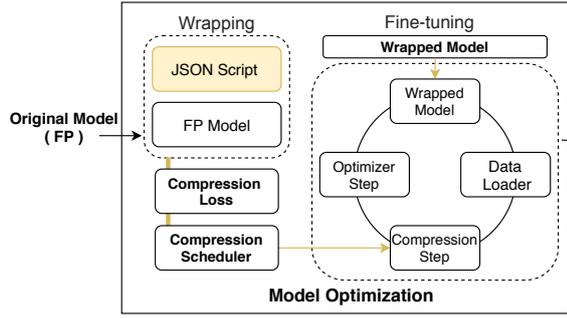
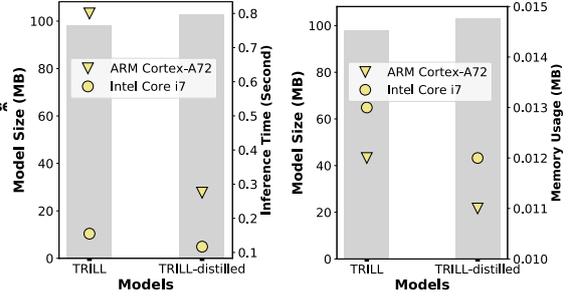

Fig. 5. The overall compression procedures: loading a JSON configuration script, then passing FP model along with the JSON file. A wrapped model then ready for compression and fine-tuning (NNCF [42]).

Fig. 6. CPU execution time required by each encoder (paralinguistics) models (left), and memory usage consumed by these models in two deployment environments (right).

For CPC-kmean clustering, we used the implementation by [57][2], which is a modified version of the CPC model. The encoder is a 5-layer 1D-convolutional network with kernel sizes of 10, 8, 4, 4, 4 and stride sizes of 5, 4, 2, 2, 2, respectively, resulting in a downsampling factor of 160; meaning that for a 16 kHz input, each feature encodes 10 ms of audio. This is followed by the autoregressive model which is a multi-layer LSTM network with the same hidden dimension as the encoder. Then, a k-means clustering module is trained on the outputs of either the final layer or a hidden layer of the autoregressive model. The clustering is done on the collection of all the output features at every time step $t$ of all the audio files in a given training set. After training the k-means clustering, each feature is then assigned to a cluster, and each audio file can then be discretized to a sequence of discrete units corresponding to the assigned clusters. The k-means training was done on the subset of LibriSpeech containing 100 hours of clean speech. For CPC-VQ and VAE-VQ, the implementation followed [79]. The CPC-VQ encoder consists of a convolutional layer (downsampling the input by a factor of 2), followed by a stack of 4 linear layers with ReLU activations and layer normalization after each layer, while the VAE-VQ encoder is a stack of 5 convolutional layers (downsamples the input by a factor of 2). Then, the encoder output is projected into a sequence of continuous latent vectors which are discretized using a VQ layer with 512 codes. For the CPC-VQ, the autoregressive model summarizes the discrete representations up to time $t$ into a context vector $c_t$. Using this context, the model is trained to predict future codes.

For the paralinguistic embedding, the target is to learn general representation (i.e., non-semantic) useful for personalization tasks and the medical domain. We follow the work of [71] and use TRILL embedding[3]. It is based on ResNetish [64], a variant of the standard ResNet-50 architecture, followed by a $d$ = 512-dimensional embedding layer. The TRILL model uses triplet loss as a training objective, often used in similarity learning, aiming to discriminate against the same/different audio segments. Intuitively, the objective is attempting to learn an embedding of the input such that positive examples end up closer to their anchors than the corresponding negatives do. The produced embedding of dimension $d$ = 512 represents the training input of downstream paralinguistics tasks.

4.2.2 *Baseline Training.* For the linguistics embedding, the training-100 set of LibriSpeech [62] is used as a training dataset. It has multiple speakers and was recorded at a sampling rate of 16 kHz. The log Mel-spectrogram context windows with $f$ = 80 Mel bands and $t$ = 96 frames representing 0.96 s of input audio are computed from the speech waveform (i.e., STFT computed with 25 ms windows with step 10 ms) as model input. For the paralinguistic embedding, a subset of AudioSet [23] is used as a training dataset, which is the largest dataset for general purpose audio machine learning (serving as an audio equivalent of ImageNet). The log Mel-spectrogram context windows with $f$ = 64 Mel bands and $t$ = 96 frames representing 0.96 s of input audio are computed from the speech waveform (i.e., STFT computed with 25 ms windows with step 10 ms) as model input.

---
[2]https://github.com/bootphon/zerospeech2021_baseline
[3]https://github.com/google-research/non_semantic_speech_benchmark

To develop an edge-friendly model, we implement various optimization techniques and show their effect on the trade-off between performance and accuracy. Optimization techniques can be applied either during or after the training. Upon completion of model training, we apply the optimization methods and fine-tune them.

**Linguistics.** To quantize the linguistics models (i.e., CPC-kmean, CPC-VQ, and VAE-VQ), we use the Neural Network Compression Framework (NNCF) [42], i.e., a framework for neural network compression with fine-tuning to experiment with different compression techniques. It supports various compression algorithms including quantization, pruning, and sparsity applied during the model fine-tuning process to achieve better compression parameters and accuracy. The overall compression procedures can be summarized as loading a JSON configuration script that contains NNCF-specific parameters determining the compression to be applied to the model, and then passing the FP model along with the configuration script to the "nncf.create_compressed_model" function. This function returns a wrapped model ready for compression and fine-tuning, and an additional object to allow further control of the compression during the fine-tuning process, as in Figure 5. Fine-tuning is a necessary step in some cases to recover the ability to generalize what may have been damaged by the model optimization techniques. We therefore fine-tune the model over 10 epochs after implementing the model optimization (i.e., quantization) to enhance its accuracy.

**Paralinguistics.** To optimize the paralinguistics model, we follow Peplinski et al.'s work [64] and use knowledge distillation to distill "TRILL" to a much smaller student model by using a truncated MobileNet architecture. Knowledge is transferred from the teacher model to the student by minimizing a loss function, aimed at matching softened teacher logits as well as ground-truth labels [30]. The logits are softened by applying a temperature scaling function in the softmax, effectively smoothing out the probability distribution and revealing inter-class relationships learned by the teacher. The MobileNet architecture uses a width multiplier alpha to control the number of filters in the convolutional layers within each inverted residual block, and thus the student models can be distilled with several values of alpha, allowing independent variation of the width (via alpha) and depth (via truncation) of the student model while sampling a wide range of parameter counts [64]. Therefore, we distill the "TRILL" embedding to a student model which is trained to map input spectrogram to the output representation produced by "TRILL". Student embeddings are then used as input representation for solving paralinguistics tasks on edge devices.

*4.2.3 Metrics.* For system performance, we consider two metrics of computational efficiency on a MacBook Pro and a Raspberry Pi 4B, (1) CPU Execution Time, measured in seconds (s), and (2) Memory Usage, measured in megabytes (MB).

By focusing on the linguistic level, we evaluate the quality of the learned embedding using linguistic metrics introduced by [43]. These metrics either are derived by computing pseudo-distance or by computing (pseudo-) probabilities. Distance-based metrics require models to provide a pseudo-distance computed over pairs of embeddings. Probability-based metrics are computed over pairs of inputs. One example is to evaluate the syntactic abilities of language models by comparing the probabilities of grammatical versus ungrammatical sentences (i.e., syntactic level), while the other is to evaluate the lexical level by comparing the pseudo-probability associated with words and non-words. They give interpretable scores at each linguistic level: phonetics (ABX score), lexicon (spot-the-word), syntax (acceptability judgment), and semantics (similarity score). We draw on evaluations presented in the Zero Resource challenge 2021 [57] and select metrics that enable us to evaluate two linguistic levels: the acoustic and the semantics levels. This allows us to evaluate the learning of linguistic representations from raw audio signals without any labels as well as without the need for the reconstruction step.

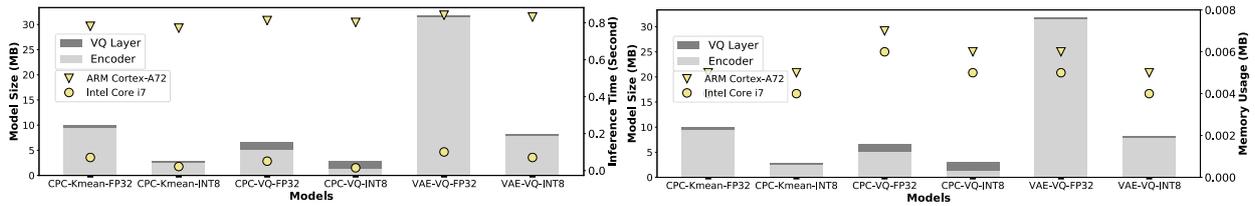

Fig. 7. CPU execution time required by each encoder (including linguistics) models (left), and memory usage consumed by these models in two deployment environments (right).

To evaluate the quality of paralinguistics embeddings, we train a set of simple models (i.e., using the Scikit-Learn library) using embeddings as input representation to solve each paralinguistics classification task. Embeddings for each utterance are averaged in time to produce a single embedding vector. We report the test accuracy (i.e., the ratio of correctly predicted labels) across combinations of downstream classifiers.

## 5 EVALUATION

In this section, we evaluate our results in terms of (i) system performance, (ii) utility evaluation, and (iii) privacy estimation.

### 5.1 System Performance

To evaluate the system performance, we measure CPU execution time and the memory usage of both baseline and compressed models during the inference time. We perform a detailed analysis of these models, each of which requires: loading the model, pre-processing the raw recordings, and producing the encoded representation. Figures 7 and 6 provide the performance analysis of two encoder types for feature disentanglement at the edge. This analysis can help us to make good decisions when choosing the appropriate optimization methods for EDGY deployment and understanding the associated cost.

*5.1.1 **Linguistics.*** We compare the performance of three different models that learn discrete linguistics representations as baseline models, namely 'CPC-Kmean-FP32', 'CPC-VQ-FP32', and 'VAE-VQ-FP32' trained using floating-point-32 precision. Firstly, we divided the models into encoder and vector quantization modules to enable better performance analysis and more distributed deployment settings. We run the pre-trained vector-quantized based models on the MacBook Pro and the Raspberry Pi 4, and we measure each module separately (i.e., encoder and quantization). As shown in Figure 7, the results indicate that we can deploy these models on the different edge/cloud devices with promising overall inference time and memory usage by all the modules (i.e., encoder and quantization) per model. For example, the 'CPC-Kmean-FP32' model requires 0.071 s and 0.001 s (i.e., encoding and quantizing) and 0.801 s and 0.025 s (i.e., encoding and quantizing) for inference time, while memory consumption is 0.005 MB and 0.004 MB (i.e., encoding and quantizing) and 0.006 MB and 0.005 MB (i.e., encoding and quantizing) on the MacBook Pro and Raspberry Pi 4, respectively. Despite the increase of the inference time consumed by these modules on the Raspberry Pi 4 compared to the MacBook Pro, the memory consumption is similar in all cases. Moreover, we show the performance results for the compressed models with INT8 quantization in Figure 7, 'FP32' represents floating-point-32 models and 'INT8' represents 8-bit-integers quantization models. Interestingly, 'INT8' model quantization was able to reduce the model size by almost 4x. We also measure the inference time and memory usage on the MacBook Pro and Raspberry Pi 4, and we find that it speeds up the inference time by about 2.4x compared with the 'FP32' baseline models.

Table 3. Overall performance of the linguistics models on evaluation datasets on two zero-shot metrics, Topline refers to the text-based language model, Baseline refers to textless representation model breaks this dependence on text, and Optimized refer to the compressed version of the baseline models.

|  |  | Topline | Baseline | | | Optimized | | |
| --- | --- | --- | --- | --- | --- | --- | --- | --- |
|  |  | BERT | CPC-Kmean | CPC - VQ | VAE - VQ | CPC-Kmean | CPC - VQ | VAE - VQ |
| Model | Set | Development Set | | | | | | |
| ABX within | Clean | 0.00 | 10.26 | 50.72 | 39.29 | 10.22 | 50.56 | 39.28 |
|  | Other | 0.00 | 14.24 | 50.22 | 42.52 | 14.16 | 50.16 | 42.47 |
| ABX across | Clean | 0.00 | 14.17 | 50.56 | 42.29 | 14.17 | 50.54 | 42.24 |
|  | Other | 0.00 | 21.44 | 50.41 | 45.13 | 21.26 | 50.24 | 45.13 |
| sSIMI | Synth. | 9.86 | 2.99 | 1.75 | 1.75 | 0.657 | 0.06 | -0.68 |
|  | Libri. | 16.11 | 6.68 | 4.42 | 2.99 | -2.75 | -5.18 | -5.25 |

*5.1.2* **Paralinguistics.** We show the performance results for two different models that learned universal representations useful for several paralinguistics tasks (e.g., speaker recognition, emotion recognition, and accent identification), namely 'TRILL' and 'TRILL-distilled' in Figure 6. We measure the inference time and memory usage on the MacBook Pro and Raspberry Pi 4. As shown in Figure 6, distillation can effectively reduce (about 2.9x) the inference time from 800 to 275 milliseconds in encoding processes on the Raspberry Pi 4. For all models, the memory usage is very small, ~12 kB. There is, however, a slight increase in the 'TRILL-distilled' model size of about 5 MB over the 'TRILL' model due to the size of its final dense layers (i.e., more parameters); see Figure 6.

**System Performance Summary.** Quantization and distillation are two techniques commonly used to address the model size and performance challenges. We found that when using these methods (i.e., quantization and knowledge distillation) we obtained a model that can be deployed on the edge with better performance in terms of inference time and memory usage. The optimized models approximately require a maximum 30 MB (*i.e.*, the linguistics layer) and 103.22 MB (i.e., the paralinguistics layer) for installation, which is suitable for limited-capacity devices. For the paralinguistics layer, however, further quantization and distillation may be needed to obtain even lighter models (*i.e.,* less than 103.22 MB). We also investigated the effect of using these methods on speeding up the inference time, achieving approximately a 2.9x improvement over the baseline models. We note that the performance of the quantized models might be varied based on the input data (*i.e.,* batch size) and the hardware used (*i.e.,* support for INT8 inference). We leave for future work additional efforts to explore these optimization approaches deployed for even more constrained devices.

## 5.2 Utility Evaluation

In our setup, edge devices disentangle the speech representations, preserve paralinguistic features, and forward the linguistic embeddings to the cloud for recognition and/or speech reconstruction. As a result, the utility of these representations is determined by the reason for learning them, whether linguistic representations are useful for speech recognition/reconstruction or not, and whether non-linguistic representations are useful for paralinguistic tasks like speaker recognition and emotion recognition in case there is a need to learn it. Consequently, we conduct two types of experiments using the representations (i.e. linguistics and paralinguistics) to assess their effectiveness in performing target-specific tasks.

*5.2.1* **Lingustics.** Linguistic representations can be evaluated at two levels, i.e., acoustic and language levels, and through two modes of operation, i.e., encoding and generation tasks. Acoustic encoding (acoustic unit discovery) involves representing speech in terms of discrete units ignoring non-linguistic components such as speaker identity and noise, while language encoding (spoken language modeling) involves learning how the language patterns are distributed [43]. Speech resynthesis involves generating audio from given acoustic units. Given that our objective is to learn linguistic representations from raw audio signals without any labels, and by following recent textless NLP approches [43], we use two automatically evaluative metrics sets: 'zero-shot' and ASR-based metrics to evaluate the quality of the learned linguistic representations to demonstrate EDGY's feasibility/compatibility with future/existing transcription systems. By using 'zero-shot' metrics [57], we analyze the quality of the encoded representations extracted from the raw recordings at two linguistic levels: phonetics (*i.e.,* how well separated phonetic categories are in a given embedding space) and semantics (*i.e.,* evaluating the semantic representation at the words level). These metrics are aimed at making it easier to evaluate discovered discrete units from raw audio without the need for any text or labeled data. We then implement a decoder to convert the encoded input units back to speech, *i.e.,* discrete units discarding non-linguistic factors, like speaker-related information. We use state-of-the-art ASR systems to translate the generated speech back to text and then apply metrics including word error rate (WER) to determine the intelligibility of the resulting speech in terms of higher linguistic content. This reflects that we have good representations extracted from the raw audio for the transcription task.

For the acoustic/phonetics level, we use the ABX score (*i.e.,* lower is better) to estimate the discriminability between phonemes. ABX is calculated by computing the distance between the representations associated with three acoustic tokens (a, b, and x), two of which belong to the same category A (a and x) and one which belongs to a different category B (b). Thus, the score is the estimated probability that a and x are closer to one another than a and b. In this paper, we use the ABX score developed in Libri-light and report both the within-speaker ABX score (where a, b and x belong to the same speaker) and the across-speaker ABX score (where a and b belong to the same speaker and x to a different one) using the development set of the Zero Resource challenge 2021. At the semantic level, the sSIMI similarity score (*i.e.,* higher is better) uses to compute the similarity of the representation of pairs of words and compares it to human similarity judgments. To obtain this, the outputs from a hidden layer of the language model of the two discretized sequences are aggregated with a pooling function to produce a fixed-length representation vector for each sequence and the cosine similarity between the two representation vectors is computed. We use the sSIMI similarity score developed in the mturk-771 dataset and report similarity distance over pairs of the same voice for the synthetic subset, and all possible pairs for the LibriSpeech subset. In this setting, the language models learn on the basis of the pseudo-text derived from clustering learned representations from scratch (without text), compared to text-based language models like BERT [57].

In Table 3, we present the results for both baseline and optimized models and compared them to topline (text-based system, namely 'BERT' [57]). The lower the ABX score, the better the linguistic representations obtained. The 'CPC-Kmean-FP32' model scored the lowest compared to the 'CPC-VQ-FP32' and 'VAE-VQ-FP32' models, in addition to a further reduction in ABX score when implementing optimization techniques by ∼0.2%. Our automatic metrics confirm the quality of the representations and outputs at the acoustic/phonetic level, but show that improvements are needed at semantic level. Although 'CPC-Kmean-FP32' scored higher compared to the 'CPC-VQ-FP32' and 'VAE-VQ-FP32' models, the overall semantic scores show the need for further improvement. The baseline here is a text-based languge model. Comparison with a text-based BERT topline system trained on the phonetic transcription shows that the speech input raises challenges for the language model component of the model that need to be addressed in further work. The language model (BERT) in this case learns on the basis of the pseudo-text derived from clustering the learned representations.

Table 4. EDGY's performance on cloud-based ASR using the word error rate (WER; lower is better) and real-time factor (RTF; lower RTF is more computationally efficient).

|         | Amazon | | Google | | IBM | | Mozilla | | Local Model | |
|---------|--------|------|--------|------|-------|------|---------|------|-------------|------|
|         | WER    | RTF  | WER    | RTF  | WER   | RTF  | WER     | RTF  | WER         | RTF  |
| Raw     | 17.26  | 57.49| 18.45  | 7.35 | 7.14  | 2.11 | 5.95    | 1.33 | 5.36        | 2.20 |
| Vocoder | 17.86  | 54.16| 63.10  | 3.46 | 72.02 | 1.11 | 47.02   | 1.39 | 8.33        | 2.19 |
| EDGY    | 23.21  | 51.41| 85.00  | 2.87 | 32.74 | 1.39 | 27.38   | 1.04 | 11.31       | 2.06 |

Finally, to demonstrate the practical applicability of EDGY with current cloud-based models (commercial Speech-to-Text APIs), we use a subset of the Librispeech test dataset (raw recordings) as a baseline, assuming that such recordings disclose all sensitive information about the user. Then, we regenerate the waveform using the pre-trained flow-based neural vocoder WaveGlow implemented in texlesslib libarary [39]. This model outputs the time-domain signal given the log Mel spectrogram as input. We also regenerated the waveform using the discovered discrete units from the raw recordings (*i.e.,* such units only maintain the linguistic information while discarding the paralinguistic/speaker-realted information). We measure the WER, which is the ratio of edit distance between words in a reference transcript and the words in the output of the speech-to-text engine to the number of words in the reference transcript (*i.e.,* lower WER means the more precise is the model), and the real-time factor (RTF), which is the ratio of CPU (processing) time to the length of the input speech file. A speech-to-text engine with lower RTF is more computationally efficient. We use the ground-truth transcripts within the dataset to calculate the WER of the raw (baseline), vocoder only, and EDGY output (privacy-aware generation). In Table 4, WER is calculated using the Automatic Speech Recognition (ASR) of current speech-to-text cloud-based services and uses the ground-truth transcripts. Examples include Amazon Transcribe [5], Google Speech [25], IBM Watson [84], Mozilla DeepSpeech [52], and a local transcription model trained on the Librispeech dataset. However, as shown in Table 4, potential discrepancies over the models' performance could be related to several factors, such as the domain mismatch between written texts and spoken utterances with ASR errors, and could also be restricted by the generation (vocoder) accuracy. The theoretical privacy vs utility comparison of these approaches is left for future work.

*5.2.2 Paralinguistics.* To support the principle of configurable privacy, and as we indicated in Sec. 2.2 that this kind of information could be useful for potential critical applications such as healthcare, we evaluate the accuracy of the paralinguistics representations for the target tasks. The overall results are in Table 5. The first three rows represent the performance of learning different paralinguistics tasks using the learned paralinguistic representation.

**Utility Evaluation Summary.** Applications such as Apple's Siri, Amazon Alexa, and others not only extract the text, but also interpret and understand the meaning so that they can respond with answers, or take action based on the user's commands. Previously, connecting an NLP application to speech inputs meant that researchers had to first train an automatic speech recognition (ASR) system, a resource-intensive operation that introduced errors, did a poor job of encoding casual linguistic interactions, and was available for just a handful of languages. Current ASR can be broken down into two stages: 1) an acoustic model that transcribes speech into text, followed by a language model to add meaning to the transcription. The audio is cut into a sequence of small chunks of 0.1 seconds, and an acoustic model tries to detect the letter pronounced in each one of them. Then, the goal of the language model is to turn them into a meaningful sequence. The performance of language models trained from text is typically evaluated using scores like perplexity. However, these systems still assume access to text to learn a language model and the mapping to the continuous audio units [81], meaning the models are only suitable for languages with very large text data sets. With textless NLP, the hope is to make ASR obsolete and to work in an end-to-end fashion, from the speech input to speech outputs. This aims at learning linguistic representations from scratch for language with little or no textual resources towards textless spoken language processing [43]. Regarding linguistic representations, we composed two zero-shot tests probing two linguistic levels: acoustic and semantic, where these metrics will help to evaluate the quality of the learned linguistic representations from raw signals (*i.e.,* unsupervised systems) without the need to reconstruct them. Therefore, these results, as shown in Table 3 are promising in learning linguistics representation directly from raw signals (*i.e.,* without text or labels) while being invariant to background noise and speaker characteristics, among others. Learning such discrete speech units might be helpful in developing more robust and inclusive speech technology, especially under limited resource-settings or where no textual resources exist for languages to be addressed in a supervised setting.

### 5.3 Privacy Estimation

Our privacy preservation objective is to achieve high ASR performance (Section 5.2.1) while maintaining low paralinguistic attribute inference performance. Specifically, the inference success of this sensitive paralinguistic information by an attacker should be less than or equal to random guessing. We employ disentangled representation learning on the speech data with the assumption that we will only provide the linguistic representations to the cloud-based services. Thus, to estimate the privacy protection, we measure to what extent linguistics representations preserved paralinguistics information. We have made our privacy estimation by using the linguistics representations and training shallow classifiers on the top of these representations. We train these small models to solve various downstream paralinguistics tasks (*i.e.,* speaker identification, emotion recognition, accent identification, and gender recognition). We report the accuracy of the classification tasks using these linguistics representations, and interestingly, based on the drop in the classification accuracy across the various paralinguistics tasks, as shown in Table 5, the discovered discrete units clearly act as an information bottleneck, forcing the acoustic models to discard speaker-related details, and shows that learning privacy-preserving linguistics representations is feasible. Our baseline approach is to train the classifiers first using the output of the paralinguistic representation learning models (full floating-point and optimized) to predict various paralinguistic tasks. For these tasks, we achieved acceptable recognition accuracy of 41.69% to 85.17%, as shown in the first three rows of Table 5. Hence, we also compare the recognition accuracy before and after quantization over acoustic features. The following describes each of them:

**Speaker Identification.**
Speaker identification involves determining which speaker has produced a given speech [53]. We use the VoxCeleb1 dataset and treat it as a multiclass classification task (i.e., a distribution over the 1,251 different speakers). As shown in Table 5, by using paralinguistics representations, we can achieve reasonable accuracy to identify the speaker. This identification accuracy decreased sharply by 14-38% when using linguistics representations (i.e., after implementing vector-quantized/clustering and optimization techniques to get these representations).

**Emotion Recognition.**
Emotion recognition involves classifying vocal emotional expressions in sentences spoken in a range of basic emotional states (e.g., happy, angry, sad, and neutral) [14]. We use CREMA-D and SAVEE datasets and consider emotion recognition as a multiclass classification task (*i.e.,* a distribution over basic emotions). We observe that the disentanglement in learning linguistics representation combined with optimization techniques shows a considerable drop in emotion recognition, *i.e.,* at a rate of 34-59 % compared with paralinguistics.

Table 5. Accuracy of downstream tasks using paralinguistics and lingustics embeddings. The top three rows refer to the performance of various paralinguistics tasks using paralinguistics representations as baseline for privacy estimation experiments. The remaining rows show performance using linguistic representations before and after quantization.

|  | Dataset → | VoxCeleb1 | | CREMA-D | | SAVEE | | Common Voice | | Common Voice | |
|---|---|---|---|---|---|---|---|---|---|---|---|
|  | Task → | Speaker | | Emotion | | Emotion | | Accent | | Gender | |
|  | Model ↓ | Eval | Test | Eval | Test | Eval | Test | Eval | Test | Eval | Test |
| Paralinguistics | TRILL - Embedding | 41.69 | 41.93 | 59.75 | 59.83 | 53.70 | 53.00 | 82.45 | 81.29 | 85.00 | 83.02 |
| | TRILL - Layer19 | 49.06 | 47.07 | 67.10 | 66.48 | 65.83 | 65.00 | 84.43 | 82.36 | 85.17 | 84.79 |
| | TRILL - Distill | 47.18 | 46.54 | 67.80 | 67.33 | 66.66 | 70.00 | 76.52 | 74.94 | 81.87 | 79.89 |
| Lingustics | **CPC** | **46.08** | **44.14** | **62.87** | **61.18** | **59.16** | **56.66** | **72.40** | **70.99** | **83.11** | **82.20** |
| | + Clustering | 17.09 | 17.00 | 29.53 | 28.85 | 20.83 | 18.33 | 23.06 | 24.39 | 44.89 | 44.41 |
| | + Clustering (QInt8) | 12.90 | 12.02 | 26.15 | 21.72 | 14.16 | 11.66 | 20.67 | 21.79 | 43.32 | 44.25 |
| | + VQ | 24.74 | 24.06 | 24.70 | 24.50 | 27.50 | 25.83 | 36.21 | 35.69 | 56.54 | 56.29 |
| | + VQ (QInt8) | 22.09 | 21.00 | 21.34 | 19.34 | 24.16 | 17.50 | 34.48 | 34.95 | 56.14 | 56.91 |
| | **VAE** | **34.67** | **34.00** | **48.60** | **46.14** | **44.16** | **35.83** | **47.44** | **45.79** | **62.13** | **60.00** |
| | + VQ | 27.90 | 27.00 | 33.19 | 31.89 | 35.83 | 35.00 | 33.04 | 31.25 | 56.19 | 56.75 |
| | + VQ (QInt8) | 21.97 | 20.09 | 27.14 | 27.00 | 29.16 | 24.16 | 31.11 | 31.11 | 55.62 | 51.50 |

**Accent/Language Identification.**
Language Identification involves classifying the language being spoken by a speaker [7]. We use the Common Voice dataset (English set only) and evaluate accent identification as a multiclass classification task (i.e., a distribution over 17 English language accents). We note that using linguistics representation combined with optimization techniques shows a significant drop in language identification, i.e., at a rate of 40-64 % compared with paralinguistics.

**Gender Recognition.**
Gender recognition involves distinguishing the speaker's gender from a given speech. We use the Common Voice dataset (English set only) and estimate gender recognition as a binary classification task (i.e., a distribution over male and female). We find that disentanglement in learning linguistics representations combined with optimization techniques can decrease gender recognition at a ratio of half compared with the learned paralinguistics representations. Interestingly, even for a binary classification task like gender classification in this case, the disentanglement approach in learning linguistics representations achieves a promising level of protection, approaching a random guess.

**Privacy Estimation Summary.** We show that these discrete speech units could be a promising solution for protecting paralinguistics information within the raw speech signals, and thus open up new areas of investigation for new privacy-aware, cross-language architectures for voice analytics systems. We used two types of representation in estimating the paralinguistics privacy, as shown in Table 5. It shows that there is a sharp drop in the classification accuracy when comparing linguistic and paralinguistics representations performance over various paralinguistics tasks (e.g., speaker recognition and emotion recognition) with an about 34% to 58% in detecting emotions for example, and interestingly, this performance drop increased by 6% when implementing optimization techniques (e.g., precision quantization). The quantization approaches show their effectiveness in learning linguistic representations, Table 3, while reducing irrelevant paralinguistics information (i.e., measured by the accuracy drop), as shown in Table 5. Such representations will help in protecting the sensitive data when sharing audio data as well as speeding up its transmission [41]. To address the configurable privacy principle, we consider also the scenario where paralinguistics information might be needed for authentication purposes or medical diagnoses, and thus we disentangle the representation learning (i.e., linguistic and paralinguistic) using a composable framework. Non-semantic and on-device embeddings can thus be tuned for privacy-sensitive applications (e.g., speaker recognition); see Table 5.

## 6 DISCUSSION AND FUTURE WORK

Preserving privacy in speech processing is still at an immature stage, and has not been adequately investigated to-date. Our experiments and findings indicate that it is possible to achieve a fair level of privacy protection at the edge while maintaining a high level of functionality for voice-controlled applications. Our results can be extended to highlight different design considerations, characterized as trade-offs, which we discuss as follows.

**Performance vs. Optimization.** First, we asked the following question: "Is it possible to develop lightweight models for representation learning that can work on the edge in near real-time while maintaining accuracy?" Deep learning model compression techniques (e.g., pruning, quantization, and knowledge distillation) aim to reduce the size of the model and its memory footprint while speeding up the inference time and saving on memory use. To do this, a pre-trained dense model is transformed into a sparse one that preserves the most important model parameters. For example, Peplinski et. al. in [64] evaluate the importance of adopting knowledge distillation to develop a set of efficient models that can learn generally-useful non-semantic speech representation and run on-device inference and training.

In our work, we have adapted several popular techniques that are currently used to compress models and enable them to be deployed at the edge. One of the primary reasons for taking an edge computing approach is to filter data locally before sending it to the cloud. Local filtering may be used to enhance the protection of users' privacy. As shown in Figure 6, combining the precision quantization with linguistics models enables us to reduce the model size by almost 4x, yielding a model about half or less the baseline 'FP32' models. Model compression is also shown to speed up inference time by about 2.4x. Interestingly, it can improve the quality of the learned linguistics representation by about 0.2% (Table 3). Moreover, having a paralinguistics embedding model trained via knowledge distillation can effectively reduce the model size and the inference time (about 2.9x). This model, in particular, is fast enough to be run in near real-time on an edge device to enable numerous privacy-sensitive applications. We were able to obtain light models for representation learning, but as future work we will seek to advance its deployment on devices with even more limited resources, e.g., by using optimization techniques like architecture reductions per module. For example, combining adversarial training [27], scalar quantization [58], and Kronecker products [76], might help to obtain even lighter models. Such models should be fast enough to run in real-time on a mobile device and present minimal performance degradation for the task(s) of interest.

**Disentanglement vs. Privacy.** Second, we examined the question: "Is it possible to increase user privacy by learning privacy-preserving representations from speech data while also increasing transparency by giving users control over the sharing of these representations?" Speech data has complex distributions and contains crucial information beyond linguistic content that may include information contained in background noise and paralinguistics information (i.e., speaker-related information), among other information. Current speech processing systems are trained without regard to the impact of these varieties of sources which may affect its effectiveness. For example, only a portion of this collected information is related to ASR, while the rest can be considered as invariant and therefore potentially impinge upon the performance of ASR systems. Likewise, the implementation of disentanglement in speaker-related representations can enhance the robustness of speaker representations and overcome common speaker recognition issues like speaker spoofing [65]. Recent studies have suggested that a disentangled speech representation can improve the interpretability and transferability of the representation in the speech signal [32]. Although such work seeks to improve the quality and effectiveness of speech processing systems, it has not considered its application to protecting privacy. We observe that learning disentangled representations can bring enhanced robustness, interpretability, and controllability. Our proposed system aims to achieve the configurable privacy principle by adopting a number of design factors. We first implement multi-layer processing based on the assumption that the information in the speech signal can be extracted via two basic layers (i.e., linguistics and paralinguistics) [32]. Second, in each layer, we distribute the processing pipeline into independent modules where each performs a specific task (i.e., mainly encoding, quantization, and classification). Finally, we apply the optimization mechanisms over the module with more computational overhead (i.e., encoding module) to enable its deployment on embedded/mobile devices. Consequently, it can be argued that the proposed system can help to develop a variety of future privacy-aware solutions between users and service providers, and can give more control to users to consent to share their data. Learning disentangled representations not only serves our purpose to protect user privacy, but is also useful in finding robust representations for different speech processing tasks with limited data in the speech domain [47, 57]. In the future, we will attempt to combine techniques like adversarial training [34] and Siamese networks [45] with disentanglement, and add further constraints grounded in information theory (e.g., triple information bottleneck [67]) to improve learning disentangled representations.

**Compactness vs. Robustness.** Finally, we investigated the question: "Is it possible to take advantage of model optimization techniques to obtain edge-friendly models as well as enhance the protection of sensitive paralinguistic information?" We first focus on the effect of compactness on privacy, and as shown in Table 5, we demonstrate that the use of compression techniques can improve filtering of sensitive information we may wish to keep locally. For example, by comparing the classification accuracy over various paralinguistics tasks using linguistics embeddings, the compressed 'INT8' models are worse than the 'FP32' baseline by about 6%. It is, therefore, also interesting to further investigate the impact of performance optimization techniques on enhancing the disentangled representations learning. Specifically, the higher degree of disentanglement between the linguistics and paralinguistics representations, the better control we have over the application of various privacy configurations.

The security vulnerabilities of DL algorithms to adversarial examples is an ongoing concern [1], but is beyond the scope of this paper. We will pose an additional future question toward enabling the deployment in a trustworthy way: "can the model compression techniques be used as a defense to improve both privacy and security objectives of voice-controlled systems?" In computer vision, for example, Gui *et al.* in [27] proposed adversarially trained model compression (ATMC) algorithm to enhance the robustness of convolutional neural networks (CNNs) against adversarial attacks that aim to fool these models into making wrong predictions. It is, therefore, also interesting to further investigate the trade-off between accuracy and optimization in addition to security. Concerning trade-offs between robustness and performance, in future work we will attempt to strengthen the representation learning models' robustness against potential attacks using advanced optimization techniques such as adversarial compression or using a combination of optimization techniques (e.g., pruning with quantization) for security and trust-sensitive voice-controlled IoT applications.

**Further Limitations.** Overall, we focused on attribute inference attacks in the speech domain, demonstrating the potential for an attacker to accurately infer a target user's sensitive or private attributes (e.g., their emotion, sex, or health status) from deep acoustic models. The goal of the privacy-preserving framework in this paper is to protect the sensitive attributes of shared data against potential attribute inference attacks launched by a curious attacker (*i.e.,* 'honest-but-curious'). We demonstrate the importance of developing privacy-preserving solutions that can run at the edge, *i.e.,* before sharing data with cloud-based services. We tested its effectiveness, however, in centralized settings with assumption of passive attack scenarios. Our next step will be to investigate whether disentanglement can be effective against active attacks in both centralized and federated learning settings, safeguarding individual users' training data. For example, Boenisch *et al.* in [11] present a new privacy attack against FL that is based on an active attacker who holds the ability to maliciously manipulate the shared model and its weights. They show how existing privacy mechanisms, such as Differentially private SGD (DPSGD), are insufficient to protect private user data from their data reconstruction attack, as current FL implementations operate under the assumption of an 'honest-but-curious' central party. While adversarial examples and overall robustness issues are out of the scope of this work, another direction of future work could be focused on extending the EDGY framework to such scenarios and demonstrating the benefits of protecting privacy while maintaining data integrity.

# 7 CONCLUSION

In this paper, we proposed EDGY, a hybrid privacy-preserving approach to delivering voice-based services that incorporates on-device paralinguistic information learning and filtering with cloud-based processing. We leverage *disentangled representation learning* to explicitly learn independent factors in the raw data. Model optimization is essential for deep learning embedded on mobile or IoT devices. Thus, we performed further combinations between the composable architecture and optimization mechanisms to accelerate deep learning inference at the edge, gaining approximately a 2.4x-2.9x performance improvement over the floating-point models. We successfully deployed our model on representative edge/cloud devices, including Raspberry Pi 4 and MacBook Pro i7, and showed its effectiveness in running in tens of milliseconds with 0.2% relative improvement in ABX score in learning linguistic representations. We also demonstrate the efficacy of our approach with current cloud-based ASR systems with minimal drop/impact on accuracy of ~5.95% WER. Using EDGY, we evaluated the trade-off made between lightweight implementation and performance, and explain that striking the correct balance will depend on the services with which we interact. We can expect further trade-offs between performance and accuracy when considering deployment for even more constrained devices.

Our future work includes investigating the development of a scheme to automatically choose a compression method and/or combine a subset of these methods so that automatic optimization can be conducted for deploying deep models for given computational resources, latency requirements, and privacy constraints.

## REFERENCES
[1] Hadi Abdullah, Kevin Warren, Vincent Bindschaedler, Nicolas Papernot, and Patrick Traynor. 2020. The Faults in our ASRs: An Overview of Attacks against Automatic Speech Recognition and Speaker Identification Systems. *arXiv preprint arXiv:2007.06622* (2020).

[2] Shimaa Ahmed, Amrita Roy Chowdhury, Kassem Fawaz, and Parmesh Ramanathan. 2020. Preech: A system for privacy-preserving speech transcription. In *29th {USENIX} Security Symposium ({USENIX} Security 20)*. 2703–2720.

[3] Ranya Aloufi, Hamed Haddadi, and David Boyle. 2019. Emotion Filtering at the Edge. In *Proceedings of the 1st Workshop on Machine Learning on Edge in Sensor Systems* (New York, NY, USA). Association for Computing Machinery. https://doi.org/10.1145/3362743.3362960

[4] Ranya Aloufi, Hamed Haddadi, and David Boyle. 2020. *Privacy-Preserving Voice Analysis via Disentangled Representations*. Association for Computing Machinery, New York, NY, USA, 1–14. https://doi.org/10.1145/3411495.3421355

[5] Amazon. 2022. Transcribe. https://aws.amazon.com/transcribe/

[6] Dario Amodei, Sundaram Ananthanarayanan, Rishita Anubhai, Jingliang Bai, Eric Battenberg, Carl Case, Jared Casper, Bryan Catanzaro, Qiang Cheng, Guoliang Chen, Jie Chen, Jingdong Chen, Zhijie Chen, Mike Chrzanowski, Adam Coates, Greg Diamos, Ke Ding, Niandong Du, Erich Elsen, Jesse Engel, Weiwei Fang, Linxi Fan, Christopher Fougner, Liang Gao, Caixia Gong, Awni Hannun, Tony Han, Lappi Vaino Johannes, Bing Jiang, Cai Ju, Billy Jun, Patrick LeGresley, Libby Lin, Junjie Liu, Yang Liu, Weigao Li, Xiangang Li, Dongpeng Ma, Sharan Narang, Andrew Ng, Sherjil Ozair, Yiping Peng, Ryan Prenger, Sheng Qian, Zongfeng Quan, Jonathan Raiman, Vinay Rao, Sanjeev Satheesh, David Seetapun, Shubho Sengupta, Kavya Srinet, Anuroop Sriram, Haiyuan Tang, Liliang Tang, Chong Wang, Jidong Wang, Kaifu Wang, Yi Wang, Zhijian Wang, Zhiqian Wang, Shuang Wu, Likai Wei, Bo Xiao, Wen Xie, Yan Xie, Dani Yogatama, Bin Yuan, Jun Zhan, and Zhenyao Zhu. 2016. Deep Speech 2: End-to-End Speech Recognition in English and Mandarin. In *Proceedings of the 33rd International Conference on International Conference on Machine Learning - Volume 48* (New

York, NY, USA). JMLR.org, 173–182.

[7] Rosana Ardila, Megan Branson, Kelly Davis, Michael Henretty, Michael Kohler, Josh Meyer, Reuben Morais, Lindsay Saunders, Francis M Tyers, and Gregor Weber. 2019. Common voice: A massively-multilingual speech corpus. *arXiv preprint arXiv:1912.06670* (2019).

[8] Alexei Baevski, Steffen Schneider, and Michael Auli. 2020. vq-wav2vec: Self-Supervised Learning of Discrete Speech Representations. In *International Conference on Learning Representations*.

[9] Alexei Baevski, Yuhao Zhou, Abdelrahman Mohamed, and Michael Auli. 2020. wav2vec 2.0: A Framework for Self-Supervised Learning of Speech Representations. *Advances in Neural Information Processing Systems* 33 (2020).

[10] Yoshua Bengio, Nicholas Léonard, and Aaron Courville. 2013. Estimating or propagating gradients through stochastic neurons for conditional computation. *arXiv preprint arXiv:1308.3432* (2013).

[11] Franziska Boenisch, Adam Dziedzic, Roei Schuster, Ali Shahin Shamsabadi, Ilia Shumailov, and Nicolas Papernot. 2021. When the Curious Abandon Honesty: Federated Learning Is Not Private. arXiv:2112.02918 [cs.LG]

[12] Antoine Boutet, Carole Frindel, Sébastien Gambs, Théo Jourdan, and Claude Rosin Ngueveu. 2020. DYSAN: Dynamically sanitizing motion sensor data against sensitive inferences through adversarial networks. *arXiv preprint arXiv:2003.10325* (2020).

[13] Cristian Buciluundefined, Rich Caruana, and Alexandru Niculescu-Mizil. 2006. Model Compression. In *Proceedings of the 12th ACM SIGKDD International Conference on Knowledge Discovery and Data Mining* (Philadelphia, PA, USA) *(KDD '06)*. Association for Computing Machinery, New York, NY, USA, 535–541. https://doi.org/10.1145/1150402.1150464

[14] Houwei Cao, David G Cooper, Michael K Keutmann, Ruben C Gur, Ani Nenkova, and Ragini Verma. 2014. Crema-d: Crowd-sourced emotional multimodal actors dataset. *IEEE transactions on affective computing* 5, 4 (2014), 377–390.

[15] William Chan, Navdeep Jaitly, Quoc Le, and Oriol Vinyals. 2016. Listen, attend and spell: A neural network for large vocabulary conversational speech recognition. In *2016 IEEE International Conference on Acoustics, Speech and Signal Processing (ICASSP)*. IEEE, 4960–4964.

[16] Ju chieh Chou, Cheng chieh Yeh, and Hung yi Lee. 2019. One-shot Voice Conversion by Separating Speaker and Content Representations with Instance Normalization. arXiv:1904.05742 [cs.LG]

[17] Ju chieh Chou, Cheng chieh Yeh, Hung yi Lee, and Lin shan Lee. 2018. Multi-target Voice Conversion without Parallel Data by Adversarially Learning Disentangled Audio Representations. arXiv:1804.02812 [eess.AS]

[18] Chung-Cheng Chiu, Tara Sainath, Yonghui Wu, Rohit Prabhavalkar, Patrick Nguyen, Zhifeng Chen, Anjuli Kannan, Ron Weiss, Kanishka Rao, Ekaterina Gonina, Navdeep Jaitly, Bo Li, Jan Chorowski, and Michiel Bacchiani. 2018. State-of-the-Art Speech Recognition with Sequence-to-Sequence Models. 4774–4778. https://doi.org/10.1109/ICASSP.2018.8462105

[19] Jan Chorowski, Ron J Weiss, Samy Bengio, and Aäron van den Oord. 2019. Unsupervised speech representation learning using wavenet autoencoders. *IEEE/ACM transactions on audio, speech, and language processing* 27, 12 (2019), 2041–2053.

[20] Nicholas Cummins, Alice Baird, and Bjoern W Schuller. 2018. Speech analysis for health: Current state-of-the-art and the increasing impact of deep learning. *Methods* 151 (2018), 41–54.

[21] Gauri Deshpande and Björn Schuller. 2020. An Overview on Audio, Signal, Speech, '|&' Language Processing for COVID-19. *arXiv preprint arXiv:2005.08579* (2020).

[22] Daniel J. Dubois, Roman Kolcun, Anna Maria Mandalari, Muhammad Talha Paracha, David Choffnes, and Hamed Haddadi. 2020. When Speakers Are All Ears: Characterizing Misactivations of IoT Smart Speakers. In *Proceedings of the 20th Privacy Enhancing Technologies Symposium (PETS 2020)* (Montreal, Canada).

[23] Jort F Gemmeke, Daniel PW Ellis, Dylan Freedman, Aren Jansen, Wade Lawrence, R Channing Moore, Manoj Plakal, and Marvin Ritter. 2017. Audio set: An ontology and human-labeled dataset for audio events. In *2017 IEEE International Conference on Acoustics, Speech and Signal Processing (ICASSP)*. IEEE, 776–780.

[24] Yuan Gong and Christian Poellabauer. 2018. Towards learning fine-grained disentangled representations from speech. *arXiv preprint arXiv:1808.02939* (2018).

[25] Google. 2022. Speech-to-Tex12t. https://cloud.google.com/speech-to-text

[26] Filip Granqvist, Matt Seigel, Rogier van Dalen, Áine Cahill, Stephen Shum, and Matthias Paulik. 2020. Improving on-device speaker verification using federated learning with privacy. *arXiv preprint arXiv:2008.02651* (2020).

[27] Shupeng Gui, Haotao N Wang, Haichuan Yang, Chen Yu, Zhangyang Wang, and Ji Liu. 2019. Model compression with adversarial robustness: A unified optimization framework. In *Advances in Neural Information Processing Systems*. 1285–1296.

[28] Sanaul Haq, Philip JB Jackson, and James Edge. 2008. Audio-visual feature selection and reduction for emotion classification. In *Proc. Int. Conf. on Auditory-Visual Speech Processing (AVSP'08), Tangalooma, Australia*.

[29] Yanzhang He, Tara N Sainath, Rohit Prabhavalkar, Ian McGraw, Raziel Alvarez, Ding Zhao, David Rybach, Anjuli Kannan, Yonghui Wu, Ruoming Pang, et al. 2019. Streaming end-to-end speech recognition for mobile devices. In *ICASSP 2019-2019 IEEE International Conference on Acoustics, Speech and Signal Processing (ICASSP)*. IEEE, 6381–6385.

[30] Geoffrey Hinton, Oriol Vinyals, and Jeff Dean. 2015. Distilling the Knowledge in a Neural Network. arXiv:1503.02531 [stat.ML]

[31] Lori L Holt and Andrew J Lotto. 2010. Speech perception as categorization. *Attention, Perception, & Psychophysics* 72, 5 (2010), 1218–1227.

[32] Wei-Ning Hsu, Yu Zhang, and James Glass. 2017. Unsupervised learning of disentangled and interpretable representations from sequential data. In *Advances in neural information processing systems*. 1878–1889.

[33] T. Hu, A. Shrivastava, O. Tuzel, and C. Dhir. 2020. Unsupervised Style and Content Separation by Minimizing Mutual Information for Speech Synthesis. In *ICASSP 2020 - 2020 IEEE International Conference on Acoustics, Speech and Signal Processing (ICASSP)*. 3267–3271.

[34] Wen-Chin Huang, Hao Luo, Hsin-Te Hwang, Chen-Chou Lo, Yu-Huai Peng, Yu Tsao, and Hsin-Min Wang. 2020. Unsupervised Representation Disentanglement Using Cross Domain Features and Adversarial Learning in Variational Autoencoder Based Voice Conversion. *IEEE Transactions on Emerging Topics in Computational Intelligence* (2020).

[35] Benoit Jacob, Skirmantas Kligys, Bo Chen, Menglong Zhu, Matthew Tang, Andrew Howard, Hartwig Adam, and Dmitry Kalenichenko. 2018. Quantization and training of neural networks for efficient integer-arithmetic-only inference. In *Proceedings of the IEEE Conference on Computer Vision and Pattern Recognition*. 2704–2713.

[36] Mimansa Jaiswal and Emily Mower Provost. 2019. Privacy enhanced multimodal neural representations for emotion recognition. *arXiv preprint arXiv:1910.13212* (2019).

[37] Huafeng Jin and Shuo Wang. 2018. Voice-based determination of physical and emotional characteristics of users.

[38] Nal Kalchbrenner, Erich Elsen, Karen Simonyan, Seb Noury, Norman Casagrande, Edward Lockhart, Florian Stimberg, Aaron van den Oord, Sander Dieleman, and Koray Kavukcuoglu. 2018. Efficient Neural Audio Synthesis. In *Proceedings of the 35th International Conference on Machine Learning (Proceedings of Machine Learning Research, Vol. 80)*, Jennifer Dy and Andreas Krause (Eds.). PMLR, 2410–2419.

[39] Eugene Kharitonov, Jade Copet, Kushal Lakhotia, Tu Anh Nguyen, Paden Tomasello, Ann Lee, Ali Elkahky, Wei-Ning Hsu, Abdelrahman Mohamed, Emmanuel Dupoux, and Yossi Adi. 2022. textless-lib: a Library for Textless Spoken Language Processing. arXiv:2202.07359 [cs.CL]

[40] Hyunjik Kim and Andriy Mnih. 2018. Disentangling by Factorising. In *International Conference on Machine Learning*. 2649–2658.

[41] W. Bastiaan Kleijn, Andrew Storus, Michael Chinen, Tom Denton, Felicia S. C. Lim, Alejandro Luebs, Jan Skoglund, and Hengchin Yeh. 2021. Generative Speech Coding with Predictive Variance Regularization. *arXiv preprint arXiv:2102.09660* (2021).

[42] Alexander Kozlov, Ivan Lazarevich, Vasily Shamporov, Nikolay Lyalyushkin, and Yury Gorbachev. 2020. Neural network compression framework for fast model inference. *arXiv preprint arXiv:2002.08679* (2020).

[43] Kushal Lakhotia, Evgeny Kharitonov, Wei-Ning Hsu, Yossi Adi, Adam Polyak, Benjamin Bolte, Tu-Anh Nguyen, Jade Copet, Alexei Baevski, Adelrahman Mohamed, and Emmanuel Dupoux. 2021. Generative Spoken Language Modeling from Raw Audio. arXiv:2102.01192 [cs.CL]

[44] B. M. Lal Srivastava, N. Vauquier, M. Sahidullah, A. Bellet, M. Tommasi, and E. Vincent. 2020. Evaluating Voice Conversion-Based Privacy Protection against Informed Attackers. In *ICASSP 2020 - 2020 IEEE International Conference on Acoustics, Speech and Signal Processing (ICASSP)*. 2802–2806. https://doi.org/10.1109/ICASSP40776.2020.9053868

[45] Petri-Johan Last, Herman A Engelbrecht, and Herman Kamper. 2020. Unsupervised feature learning for speech using correspondence and Siamese networks. *IEEE Signal Processing Letters* (2020).

[46] S. Latif, R. Rana, S. Khalifa, R. Jurdak, J. Epps, and B. W. Schuller. 2020. Multi-Task Semi-Supervised Adversarial Autoencoding for Speech Emotion Recognition. *IEEE Transactions on Affective Computing* (2020), 1–1.

[47] Siddique Latif, Rajib Rana, Sara Khalifa, Raja Jurdak, Junaid Qadir, and Björn W Schuller. 2020. Deep Representation Learning in Speech Processing: Challenges, Recent Advances, and Future Trends. *arXiv preprint arXiv:2001.00378* (2020).

[48] Shaoshi Ling and Yuzong Liu. 2020. DeCoAR 2.0: Deep Contextualized Acoustic Representations with Vector Quantization. *arXiv preprint arXiv:2012.06659* (2020).

[49] Mohammad Malekzadeh, Richard G Clegg, Andrea Cavallaro, and Hamed Haddadi. 2019. Mobile sensor data anonymization. In *Proceedings of the International Conference on Internet of Things Design and Implementation*. 49–58.

[50] Mohammad Malekzadeh, Richard G Clegg, Andrea Cavallaro, and Hamed Haddadi. 2020. Privacy and utility preserving sensor-data transformations. *Pervasive and Mobile Computing* (2020), 101132.

[51] Mohammad Malekzadeh, Richard G Clegg, and Hamed Haddadi. 2017. Replacement autoencoder: A privacy-preserving algorithm for sensory data analysis. *arXiv preprint arXiv:1710.06564* (2017).

[52] Mozilla. 2022. Watson Speech to Text. https://github.com/mozilla/DeepSpeech

[53] Arsha Nagrani, Joon Son Chung, and Andrew Zisserman. 2017. VoxCeleb: A Large-Scale Speaker Identification Dataset. In *Interspeech 2017, 18th Annual Conference of the International Speech Communication Association, Stockholm, Sweden, August 20-24, 2017*, Francisco Lacerda (Ed.). ISCA, 2616–2620.

[54] Vikramjit Mitra Sue Booker Erik Marchi David Scott Farrar Ute Dorothea Peitz Bridget Cheng Ermine Teves Anuj Mehta Devang Naik. 2019. Leveraging Acoustic Cues and Paralinguistic Embeddings to Detect Expression from Voice. https://arxiv.org/pdf/1907.00112.pdf

[55] Sharan Narang, Erich Elsen, Gregory Diamos, and Shubho Sengupta. 2017. Exploring sparsity in recurrent neural networks. *arXiv preprint arXiv:1704.05119* (2017).

[56] Andreas Nautsch, Abelino Jiménez, Amos Treiber, Jascha Kolberg, Catherine Jasserand, Els Kindt, Héctor Delgado, Massimiliano Todisco, Mohamed Amine Hmani, Aymen Mtibaa, Mohammed Ahmed Abdelraheem, Alberto Abad, Francisco Teixeira, Driss Matrouf, Marta Gomez-Barrero, Dijana Petrovska-Delacrétaz, Gérard Chollet, Nicholas W. D. Evans, and Christoph Busch. 2019. Preserving privacy in speaker and speech characterisation. *Comput. Speech Lang.* 58 (2019), 441–480.

[57] Tu Anh Nguyen, Maureen de Seyssel, Patricia Rozé, Morgane Rivière, Evgeny Kharitonov, Alexei Baevski, Ewan Dunbar, and Emmanuel Dupoux. 2020. The Zero Resource Speech Benchmark 2021: Metrics and baselines for unsupervised spoken language modeling. *arXiv preprint arXiv:2011.11588* (2020).

[58] Deniz Oktay, Johannes Ballé, Saurabh Singh, and Abhinav Shrivastava. 2020. Scalable Model Compression by Entropy Penalized Reparameterization. In *International Conference on Learning Representations*. https://openreview.net/forum?id=HkgxW0EYDS

[59] Aaron van den Oord, Sander Dieleman, Heiga Zen, Karen Simonyan, Oriol Vinyals, Alex Graves, Nal Kalchbrenner, Andrew Senior, and Koray Kavukcuoglu. 2016. Wavenet: A generative model for raw audio. *arXiv preprint arXiv:1609.03499* (2016).

[60] Aaron van den Oord, Yazhe Li, and Oriol Vinyals. 2018. Representation learning with contrastive predictive coding. *arXiv preprint arXiv:1807.03748* (2018).

[61] Seyed Ali Osia, Ali Shahin Shamsabadi, Sina Sajadmanesh, Ali Taheri, Kleomenis Katevas, Hamid R. Rabiee, Nicholas D. Lane, and Hamed Haddadi. 2020. A Hybrid Deep Learning Architecture for Privacy-Preserving Mobile Analytics. *IEEE Internet of Things Journal* 7, 5 (2020), 4505–4518.

[62] Vassil Panayotov, Guoguo Chen, Daniel Povey, and Sanjeev Khudanpur. 2015. Librispeech: an asr corpus based on public domain audio books. In *2015 IEEE International Conference on Acoustics, Speech and Signal Processing (ICASSP)*. IEEE, 5206–5210.

[63] Jong-Hyeon Park, Myungwoo Oh, and Hyung-Min Park. 2019. Unsupervised Speech Domain Adaptation Based on Disentangled Representation Learning for Robust Speech Recognition. *arXiv preprint arXiv:1904.06086* (2019).

[64] Jacob Peplinski, Joel Shor, Sachin Joglekar, Jake Garrison, and Shwetak Patel. 2020. FUN! Fast, Universal, Non-Semantic Speech Embeddings. arXiv:2011.04609 [cs.SD]

[65] Raghuveer Peri, Haoqi Li, Krishna Somandepalli, Arindam Jati, and Shrikanth Narayanan. 2020. An empirical analysis of information encoded in disentangled neural speaker representations. In *Proceedings of Odyssey* (Tokyo, Japan).

[66] Jianwei Qian, Haohua Du, Jiahui Hou, Linlin Chen, Taeho Jung, and Xiang-Yang Li. 2018. Hidebehind: Enjoy Voice Input with Voiceprint Unclonability and Anonymity. In *Proceedings of the 16th ACM Conference on Embedded Networked Sensor Systems* (Shenzhen, China). Association for Computing Machinery, 82–94. https://doi.org/10.1145/3274783.3274855

[67] Kaizhi Qian, Yang Zhang, Shiyu Chang, David Cox, and Mark Hasegawa-Johnson. 2020. Unsupervised speech decomposition via triple information bottleneck. *arXiv preprint arXiv:2004.11284* (2020).

[68] Nisarg Raval, Ashwin Machanavajjhala, and Jerry Pan. 2019. Olympus: sensor privacy through utility aware obfuscation. *Proceedings on Privacy Enhancing Technologies* 2019, 1 (2019), 5–25.

[69] Morgane Rivière, Armand Joulin, Pierre-Emmanuel Mazaré, and Emmanuel Dupoux. 2020. Unsupervised pretraining transfers well across languages. In *ICASSP 2020-2020 IEEE International Conference on Acoustics, Speech and Signal Processing (ICASSP)*. IEEE, 7414–7418.

[70] Björn Schuller, Stefan Steidl, Anton Batliner, Felix Burkhardt, Laurence Devillers, Christian MüLler, and Shrikanth Narayanan. 2013. Paralinguistics in speech and language—state-of-the-art and the challenge. *Computer Speech & Language* 27, 1 (2013), 4–39.

[71] Joel Shor, Aren Jansen, Ronnie Maor, Oran Lang, Omry Tuval, Felix de Chaumont Quitry, Marco Tagliasacchi, Ira Shavitt, Dotan Emanuel, and Yinnon Haviv. 2020. Towards Learning a Universal Non-Semantic Representation of Speech. *arXiv preprint arXiv:2002.12764* (2020).

[72] Congzheng Song and Vitaly Shmatikov. 2020. Overlearning Reveals Sensitive Attributes. In *International Conference on Learning Representations*.

[73] Brij Mohan Lal Srivastava, Aurélien Bellet, Marc Tommasi, and Emmanuel Vincent. 2019. Privacy-preserving adversarial representation learning in ASR: Reality or illusion? *arXiv preprint arXiv:1911.04913* (2019).

[74] Brij Mohan Lal Srivastava, Natalia Tomashenko, Xin Wang, Emmanuel Vincent, Junichi Yamagishi, Mohamed Maouche, Aurélien Bellet, and Marc Tommasi. 2020. Design choices for x-vector based speaker anonymization. *arXiv preprint arXiv:2005.08601* (2020).

[75] G. Sun, Y. Zhang, R. J. Weiss, Y. Cao, H. Zen, and Y. Wu. 2020. Fully-Hierarchical Fine-Grained Prosody Modeling For Interpretable Speech Synthesis. In *ICASSP 2020 - 2020 IEEE International Conference on Acoustics, Speech and Signal Processing (ICASSP)*. 6264–6268.

[76] Urmish Thakker, Jesse Beu, Dibakar Gope, Chu Zhou, Igor Fedorov, Ganesh Dasika, and Matthew Mattina. 2019. Compressing rnns for iot devices by 15-38x using kronecker products. *arXiv preprint arXiv:1906.02876* (2019).

[77] Natalia Tomashenko, Brij Mohan Lal Srivastava, Xin Wang, Emmanuel Vincent, Andreas Nautsch, Junichi Yamagishi, Nicholas Evans, Jose Patino, Jean-François Bonastre, Paul-Gauthier Noé, et al. 2020. Introducing the VoicePrivacy initiative. *arXiv preprint arXiv:2005.01387* (2020).

[78] Aaron van den Oord, Oriol Vinyals, et al. 2017. Neural discrete representation learning. In *Advances in Neural Information Processing Systems*. 6306–6315.

[79] Benjamin van Niekerk, Leanne Nortje, and Herman Kamper. 2020. Vector-quantized neural networks for acoustic unit discovery in the ZeroSpeech 2020 challenge. *arXiv preprint arXiv:2005.09409* (2020).

[80] Lisa van Staden and Herman Kamper. 2020. A comparison of self-supervised speech representations as input features for unsupervised acoustic word embeddings. arXiv:2012.07387 [cs.CL]

[81] Dong Wang, Xiaodong Wang, and Shaohe Lv. 2019. An overview of end-to-end automatic speech recognition. *Symmetry* 11, 8 (2019), 1018.

[82] Xiaofei Wang, Yiwen Han, Victor CM Leung, Dusit Niyato, Xueqiang Yan, and Xu Chen. 2020. Convergence of edge computing and deep learning: A comprehensive survey. *IEEE Communications Surveys & Tutorials* 22, 2 (2020), 869–904.

[83] Y. Wang, X. Fan, I. Chen, Y. Liu, T. Chen, and B. Hoffmeister. 2019. End-to-end Anchored Speech Recognition. In *ICASSP 2019 - 2019 IEEE International Conference on Acoustics, Speech and Signal Processing (ICASSP)*.

[84] IBM Watson. 2022. Watson Speech to Text. https://www.ibm.com/ca-en/cloud/watson-speech-to-text

[85] Weidi Xie, Arsha Nagrani, Joon Son Chung, and Andrew Zisserman. 2019. Utterance-level aggregation for speaker recognition in the wild. In *ICASSP 2019-2019 IEEE International Conference on Acoustics, Speech and Signal Processing (ICASSP)*. IEEE, 5791–5795.

[86] Samuel Yeom, Irene Giacomelli, Matt Fredrikson, and Somesh Jha. 2018. Privacy risk in machine learning: Analyzing the connection to overfitting. In *2018 IEEE 31st Computer Security Foundations Symposium (CSF)*. IEEE, 268–282.

[87] Bohan Zhai, Tianren Gao, Flora Xue, Daniel Rothchild, Bichen Wu, Joseph E Gonzalez, and Kurt Keutzer. 2020. SqueezeWave: Extremely Lightweight Vocoders for On-device Speech Synthesis. *arXiv preprint arXiv:2001.05685* (2020).

[88] Hanbin Zhang, Chen Song, Aosen Wang, Chenhan Xu, Dongmei Li, and Wenyao Xu. 2019. Pdvocal: Towards privacy-preserving parkinson's disease detection using non-speech body sounds. In *The 25th Annual International Conference on Mobile Computing and Networking*. 1–16.

[89] Y. Zhang, S. Pan, L. He, and Z. Ling. 2019. Learning Latent Representations for Style Control and Transfer in End-to-end Speech Synthesis. In *ICASSP 2019 - 2019 IEEE International Conference on Acoustics, Speech and Signal Processing (ICASSP)*. 6945–6949.